\documentclass[runningheads]{llncs}

\usepackage[mobile]{eccv}

\usepackage{eccvabbrv}

\usepackage{graphicx}
\usepackage{booktabs}
\usepackage{tikz}
\usepackage{xcolor}
\usepackage{amsmath}

\usepackage[accsupp]{axessibility}  

\usepackage{hyperref}

\usepackage{orcidlink}

\definecolor{bordercolor}{HTML}{006400}
\newcommand {\mfd}[1]{\textcolor{blue}{#1}}

\begin{document}

\title{Exploiting Boundary Loss for the Hierarchical Panoptic Segmentation of Plants and Leaves} 

\titlerunning{Boundary Loss for Hierarchical Panoptic Segmentation}

\author{Madeleine Darbyshire\inst{1}\orcidlink{0000-0002-6485-6014} \and
Elizabeth Sklar\inst{2}\orcidlink{0000-0002-6383-9407} \and
Simon Parsons\inst{1}\orcidlink{0000-0002-8425-9065}}

\authorrunning{M.~Darbyshire et al.}

\institute{Lincoln Centre for Autonomous Systems (LCAS) \\
University of Lincoln, Brayford Pool, Lincoln \\
\email{25696989@students.lincoln.ac.uk}, \email{sparsons@lincoln.ac.uk} \and
Lincoln Institute of Agri-Food Technology \\
University of Lincoln, Riseholme Hall, Lincoln \\
\email{esklar@lincoln.ac.uk}}

\maketitle

\begin{abstract}
  Precision agriculture leverages data and machine learning so that farmers can monitor their crops and target interventions precisely. This enables the precision application of herbicide only to weeds, or the precision application of fertilizer only to undernourished crops, rather than to the entire field. The approach promises to maximize yields while minimizing resource use and harm to the surrounding environment. To this end, we propose a hierarchical panoptic segmentation method that simultaneously determines leaf count (as an identifier of plant growth) and locates weeds within an image. In particular, our approach aims to improve the segmentation of smaller instances like the leaves and weeds by incorporating focal loss and boundary loss. Not only does this result in competitive performance, achieving a \(PQ^\dagger\) of 81.89 on the standard training set, but we also demonstrate we can improve leaf-counting accuracy with our method. The code is available at \url{https://github.com/madeleinedarbyshire/HierarchicalMask2Former}.
  \keywords{Precision Agriculture \and Segmentation}
\end{abstract}

\section{Introduction}
\label{sec:intro}
The continued growth of the global population has put farmers under pressure to produce more food to meet the increasing demand. However, concerns about the environmental impact of agriculture are placing simultaneous pressure on farmers to mitigate the environmental harms of their operations. All the while, climate change is making growing conditions more unpredictable leading to new challenges in providing a reliable food supply. 

\emph{Precision agriculture} aims to leverage data and machine learning to help farmers make informed decisions, and target interventions precisely. For example, herbicide usage can be reduced by first detecting and then only targeting weeds rather than spraying the entire field with herbicide. Furthermore, crop monitoring can indicate where fertilizer should be targeted for healthy plant growth. Various phenotypic traits can be used as indicators of crop growth, but in this paper, we use leaf count.

The paper aims to combine crop and weed segmentation as well as leaf segmentation in a single hierarchical panoptic segmentation architecture. The task is referred to as hierarchical because of the hierarchical structure of plants. Our approach adapts the latest state-of-the-art panoptic segmentation architecture to the task and incorporates additional loss functions to tune it to better segment small areas such as weeds and leaves.

\begin{figure}[t]
\centering
\begin{subfigure}{0.3\columnwidth}
    \centering
    \includegraphics[width=0.95\linewidth]{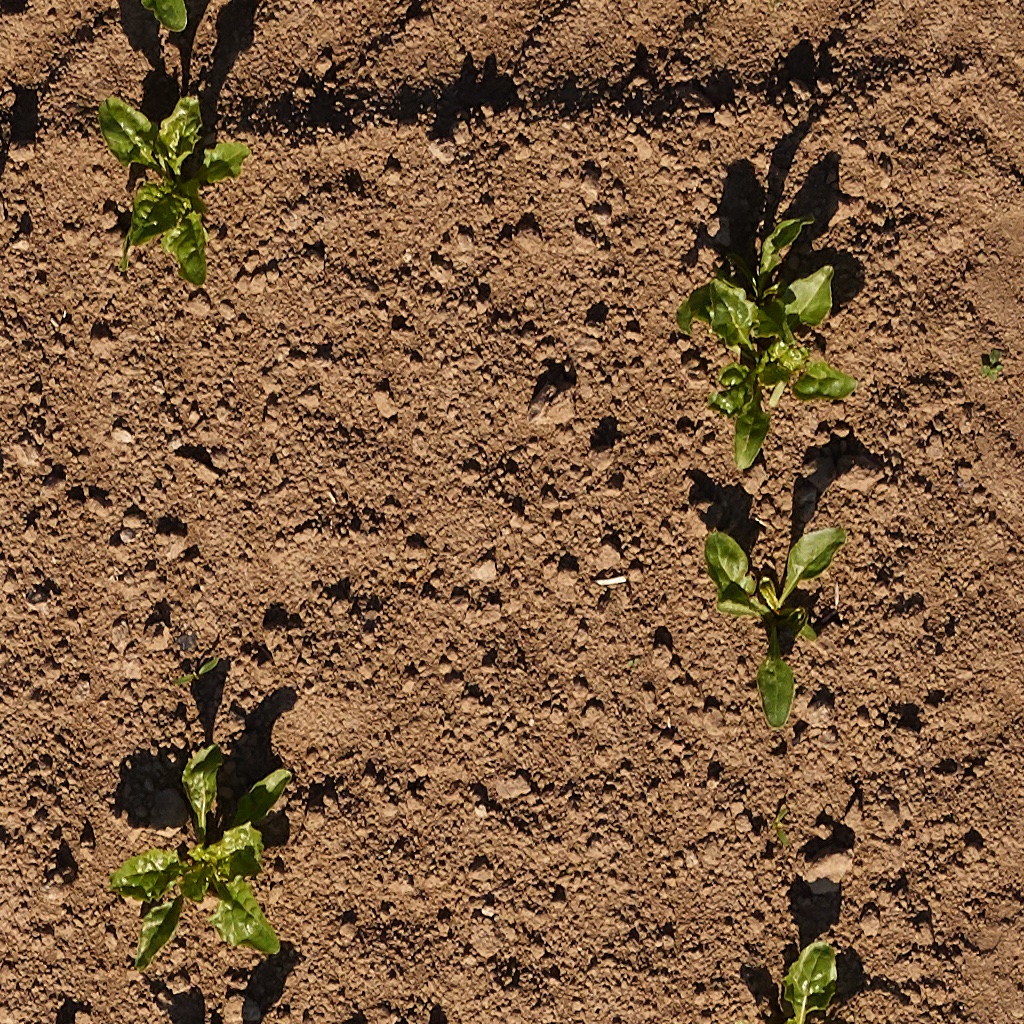}\vspace{0.3cm}
    \includegraphics[width=0.95\linewidth]{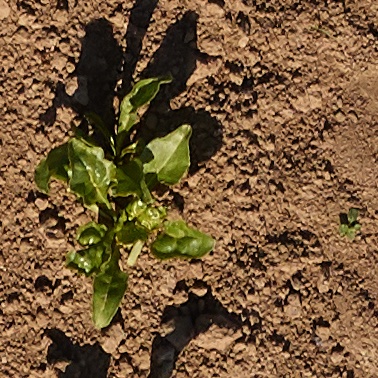}
    \caption{Original}
\end{subfigure}
\begin{subfigure}{0.3\columnwidth}
    \centering
    \includegraphics[width=0.95\linewidth]{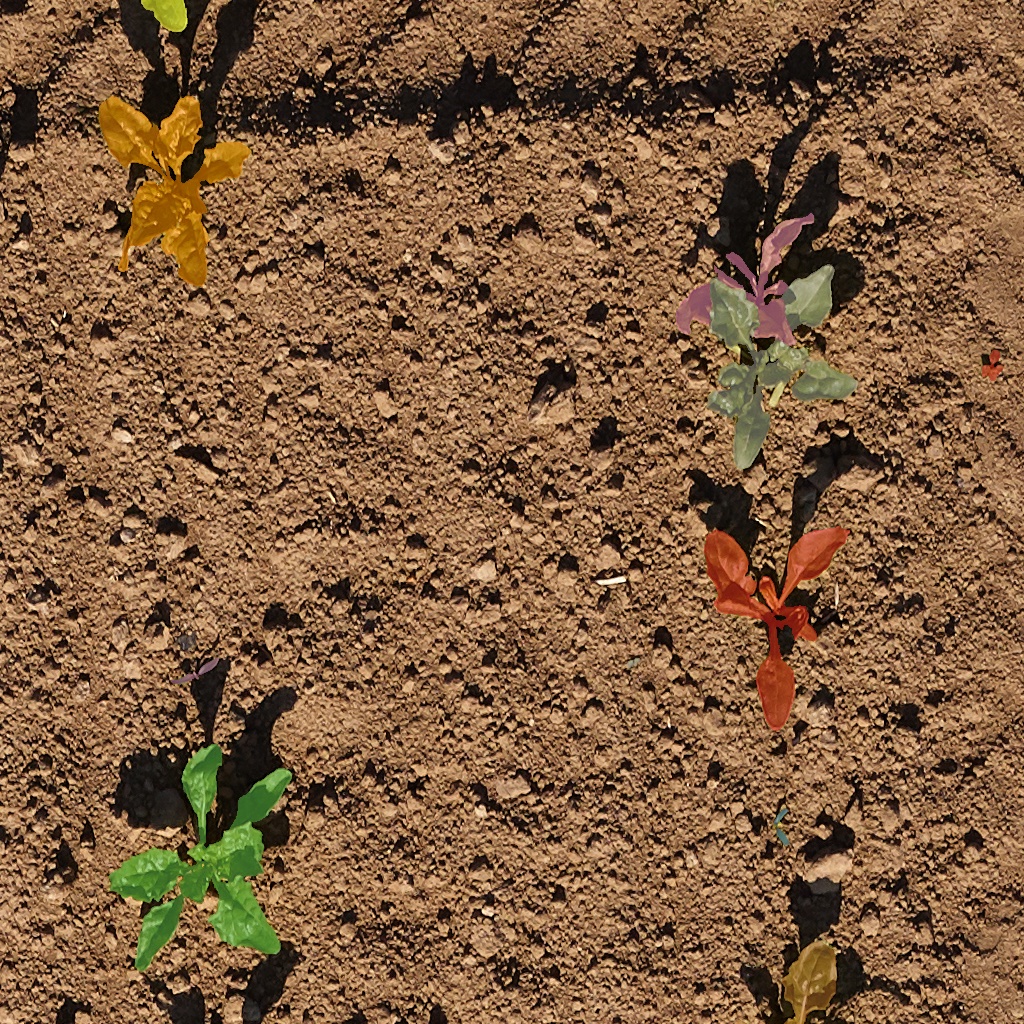}\vspace{0.3cm}
    \includegraphics[width=0.95\linewidth]{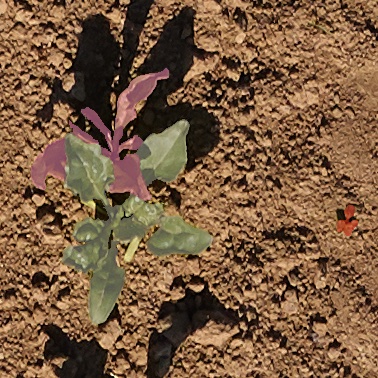}
    \caption{Plant Instances}
\end{subfigure}
\begin{subfigure}{0.3\columnwidth}
    \centering
    \includegraphics[width=0.95\linewidth]{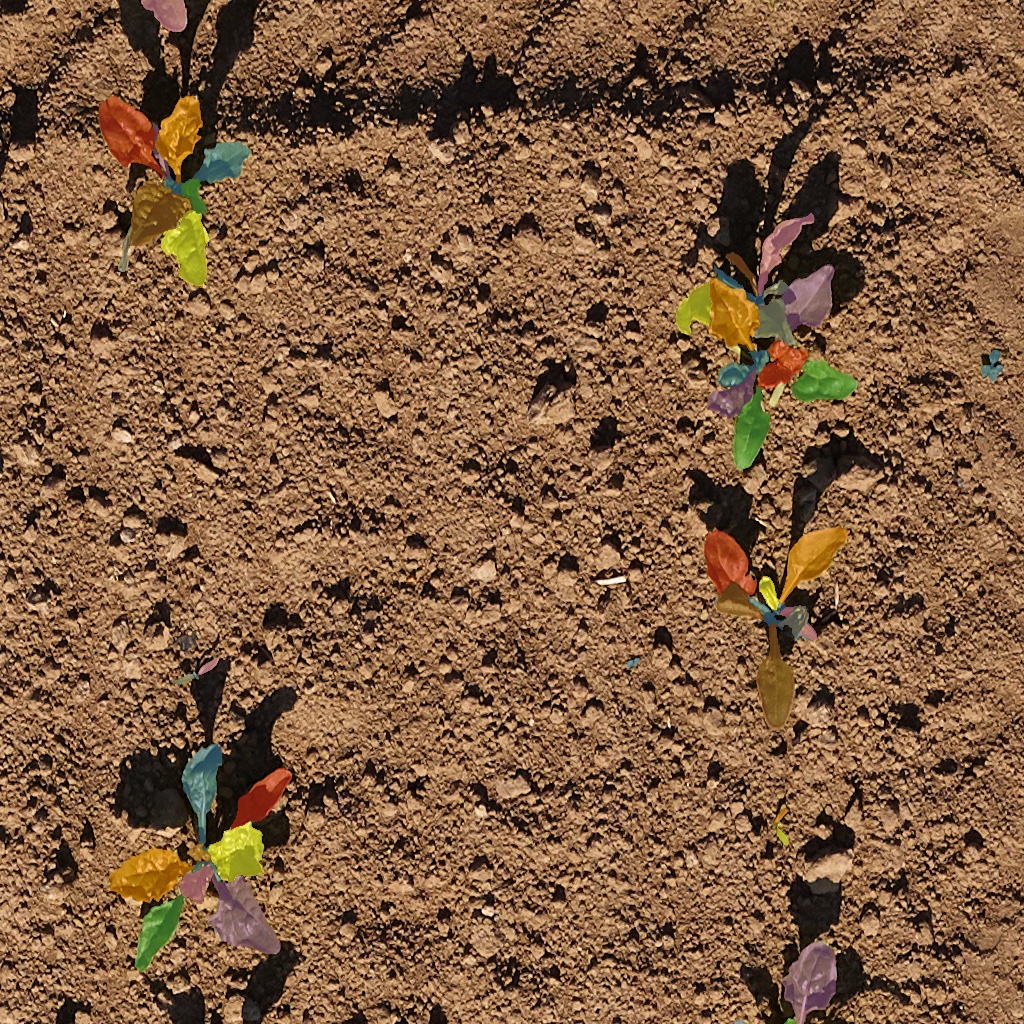}\vspace{0.3cm}
    \includegraphics[width=0.95\linewidth]{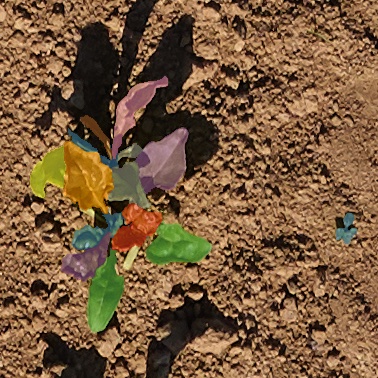}
    \caption{Leaf Instances}
\end{subfigure}
\caption{Example images from the PhenoBench dataset. On the top are the full size images and the bottom shows a close-up of the same image. Images (a) are the originals, (b) show whole plants segmented at the pixel level, and (c) show individual leaves segmented at the pixel level.}
\label{fig:examples}
\end{figure}

\section{Related Work}

\subsubsection{Semantic Segmentation}
In recent years, most approaches to crop and weed segmentation have utilised deep learning in one form or another. Early examples used SegNet \cite{badrinarayanan2017segnet} to classify each pixel in the image as either crop, weed or background \cite{di2017automatic, sa2018weednet}.  SegNet employs an encoder-decoder structure, where the encoder extracts hierarchical features from input images, and the decoder produces pixel-wise segmentation masks.  This encoder-decoder segmentation architecture was further improved in DeepLabV3+ \cite{chen2017rethinking} which added atrous convolutions to capture larger spatial context. Moreover, U-Net \cite{ronneberger2015u}, another similar segmentation approach, introduces skip connections that allow it to capture both high-level and low-level features. U-Net was shown to outperform DeepLabV3+ in crop and weed segmentation \cite{genze2022deep, zou2021modified}. 

\subsubsection{Instance Segmentation}
Beyond classifying the pixels, instance segmentation has been used to distinguish between individual crop and weed plants. Mask R-CNN \cite{osorio2020deep}, a widely used instance segmentation technique, combines pixel-level semantic segmentation with object bounding box predictions to segment instances. This approach can be applied to many panoptic segmentation problems as well. Panoptic-DeepLab \cite{cheng2020panoptic} builds on an adapted version of DeepLabV3+, adding segmentation heads to make it suitable for instance segmentation and panoptic segmentation. Mask2Former \cite{cheng2021mask2former}, and its predecessor MaskFormer \cite{cheng2021maskformer}, utilise an approach to instance and panoptic segmentation that differs from these per-pixel approaches. Instead, images are partitioned into a number of regions, represented with binary masks, then each of these is assigned a class. This approach achieved state-of-the-art performance on benchmark datasets \cite{cheng2021mask2former}.

\subsubsection{Segmentation of Crops and Leaves}
Adjacent to the problem of identifying individual plant instances is the identification of individual leaf instances within each plant instance. A catalyst for research in this area was the CVPPP Leaf Segmentation Challenge \cite{scharr2014annotated}. The goal of this was to extract plant traits from images of single plants in laboratory conditions. Aich et el. suggested a successful approach was where SegNet was used to create binary masks of the plants and a regression network was used to count the leaves \cite{aich2017leaf}. More recently, Weyler et al. demonstrated each plant and its leaves could be identified from images containing multiple plants taken under real field conditions \cite{weyler2022joint, weyler2022field}. 

Subsequently, the tasks of crop and weed segmentation and leaf instance segmentation were combined in the HAPT architecture \cite{roggiolani2023hierarchical} where ERFNet \cite{romera2017erfnet} is adapted and a second decoder is added so that there is one decoder for generating plant masks and another for generating leaf masks. Subsequently, a competition at CVPPA@ICCV 2023 \cite{cvppa} showcased state-of-the-art approaches such as Mask2Former \cite{cheng2021mask2former} and SAM \cite{kirillov2023segment} on the Phenobench crop, leaf and weed dataset \cite{weyler2023dataset}, shown in Figure \ref{fig:examples}. These approaches, in addition to HAPT, demonstrated strong performance on crop segmentation but performance was worse on leaves and weeds. 

\subsubsection{Segmentation of Small Areas}
One issue that arises when applying standard state-of-the-art approaches to segment weeds and leaves is that they tend to use loss functions that perform best on more balanced segmentation problems. While it is often the case in segmentation problems that there are more background pixels than foreground pixels in a segmentation mask, the masks of weeds and some leaves are highly unbalanced with very few foreground pixels. Regional integrals such as Dice and cross-entropy are commonly used, but on highly unbalanced segmentations, can easily be biased towards the majority class. Alternatives to these loss functions, addressing the challenge of highly unbalanced segmentation have often come from the medical applications where these significant imbalances are common \cite{ma2021loss}. Kervadec et al. \cite{kervadec2019boundary} proposed a boundary loss function that uses integrals over the interface between regions instead of unbalanced integrals over the regions. This proposes a potential solution to the imbalance we want to address.

In this work, we demonstrate how implementing the focal loss \cite{lin2017focal} and boundary loss \cite{kervadec2019boundary} can improve segmentation performance. Importantly, for the real-world application of this technology, these can be employed to improve relatively lightweight architectures without having an impact on model footprint or speed at inference time. Additionally, we evaluate how accurate the leaf counts are. Finally, we show that our approach is competitive with the state-of-the-art.

\section{Methodology}
\subsection{Adapted Mask2Former Architecture}
\begin{figure}[tb]
  \centering
  \includegraphics[width=0.95\linewidth]{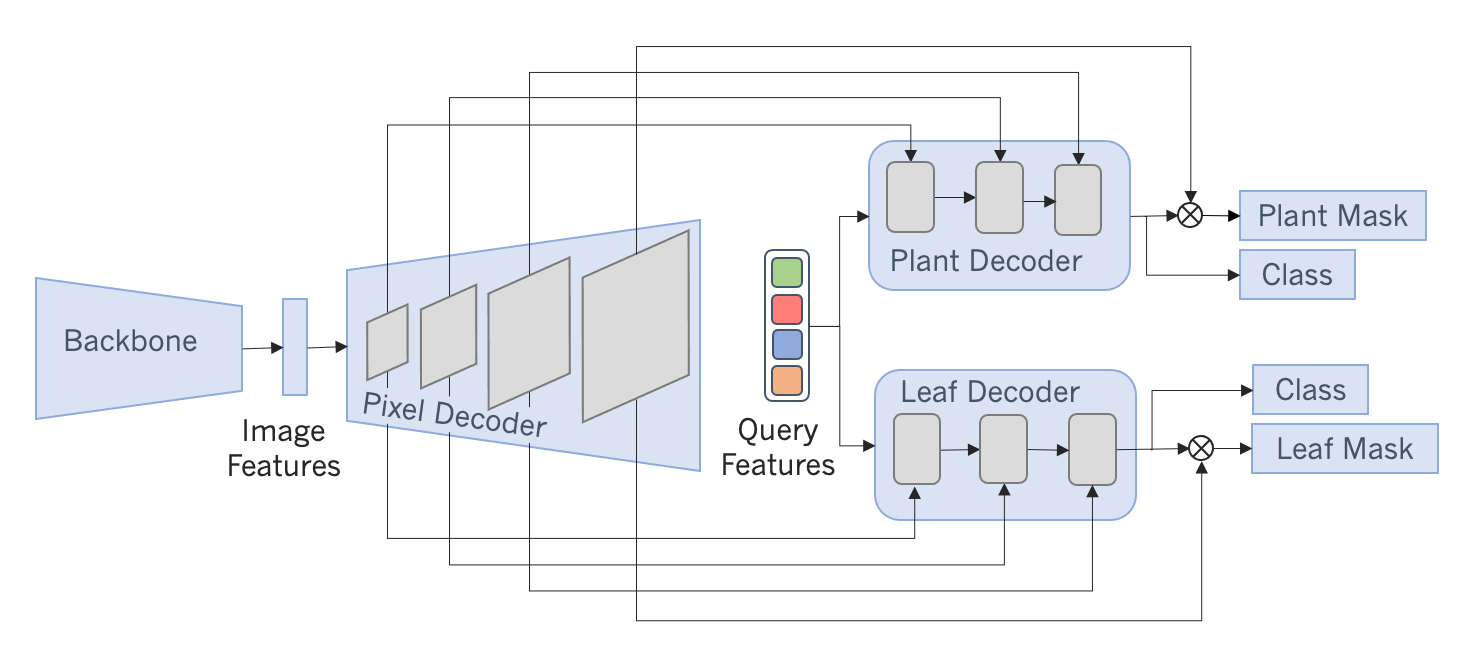}
  \caption{Adapted Mask2Former architecture with an additional transformer decoder for segmenting leaves.}
  \label{fig:lobster}
\end{figure}

We adapted the Mask2Former architecture for hierarchical panoptic segmentation due to its state-of-the-art performance on benchmark datasets in panoptic segmentation. Specifically, we enhanced the original Mask2Former by integrating an additional transformer decoder, as depicted in Figure \ref{fig:lobster}. This modification allows the architecture to generate both plant and leaf masks simultaneously. Below, we detail each architectural component and the specific adjustments made to incorporate this additional transformer.

\textbf{Backbone.} The backbone extracts low-level image features $ F^{{C_F} \times \frac{H}{S} \times \frac{W}{S}} \in \mathbb{R} $ from an input image of size $H\times W$, where $C_F$ is the number of channels and $S$ is the stride.

\textbf{Pixel decoder.} The pixel decoder gradually upsamples the low-level features to produce a feature pyramid with layers that are of resolution 1/32, 1/16 and 1/8. 
At each stage in the upsampling process, a per-pixel embedding is created $\epsilon_{pixel} \in \mathbb{R}^{C_{\epsilon} \times H \times W}$, where $C_{\epsilon}$ is the embedding dimension. In this implementation, the advanced multi-scale deformable attention Transformer, \emph{MSDeformAttn} \cite{zhu2020deformable} is used as the pixel decoder. 

\textbf{Transformer decoders.} While the standard Mask2Former model has a transformer decoder consisting of 3 transformer layers for each layer of the feature pyramid, we found no reduction in performance even with just 1 transformer layer for each layer of the feature pyramid. Given there are 3 layers in the feature pyramid, our transformer decoder has just 3 transformer decoder layers. 

Each transformer decoder layer consists of a self-attention layer, a cross-attention layer and a feed-forward network. Query features are associating with the positional embeddings produced by the pixel decoder. These query features are learnable and thus updated by each layer of the network. The transformer outputs $N$ per-segment embeddings, $Q^{C_Q \times N} \in \mathbb{R}$, where $N$ is the number of queries and $C_Q$ is the dimension that encodes global information about the segment. 

We have two separate transformer decoders, one for plants and one for leaves. Each transformer decoder takes the same set of learnable query features and each produces N per-segment embeddings $Q_{plant}$ and $Q_{leaf}$, respectively. The output from the each transformer decoder is then converted to class predictions and masks via the separate segmentation modules.

\textbf{Segmentation Module}
The segmentation modules transform the output of their respective transformer $Q$ into masks and class predictions. To acquire class probability predictions $\{p_i \in \Delta^{K}\}^N_{i=1}$, a linear classifier and softmax activation are applied to the output. There is an additional no-object class which applies where the embedding does not correspond to any region. To generate the masks, a multi-layer perceptron converts the per-segment embeddings from the transformer into $N$ mask embeddings $\epsilon_{mask} \in \mathbb{R}^{C_\epsilon \times N}$. Lastly, binary masks $m_i \in [0, 1]^{H \times W}$ are formed via the dot product of the mask embeddings, $\epsilon_{mask}$, and per-pixel embeddings, $\epsilon_{pixel}$, followed by a sigmoid activation $m_i[h,w] = sigmoid(\epsilon_{mask}[:, i]^T \cdot \epsilon_{pixel}[:,h,w])$. 

\subsection{Loss}
Due to the small size of weeds and leaves, the binary segmentation masks on which loss is calculated have vastly more background pixels than foreground pixels. Typical loss functions like Dice and cross-entropy are based on regional integrals taken from summations over the segmentation regions of differentiable functions \cite{kervadec2019boundary}. Where foreground pixels are outnumbered by background pixels significantly, the optimizer might prioritize reducing the loss from the background, due to its larger contribution, at the expense of the foreground.

\subsubsection{Point Sampling}
Point sampling is used in the original Mask2Former model \cite{cheng2021mask2former}. This involves calculating Dice and cross entropy loss against a sample of the most uncertain points in each mask. While this is proposed to improve training efficiency, it has the potential to correct some of this imbalance, in so far as we can assume foreground pixels are more likely than background pixels to be sampled due to higher uncertainty. However, we noticed a small performance improvement when using the nearest interpolation mode instead of the bilinear mode specified in the original source code.

\subsubsection{Focal Loss}
After sampling the points, the original Mask2Former model utilizes cross entropy loss and dice loss. However, using focal loss \cite{lin2017focal} instead may improve on simple cross-entropy loss for unbalanced segmentation by weighting difficult-to-classify pixels from the minority class more heavily and enabling them to make a greater contribution to the loss. It does this by introducing a modulating factor to the standard cross-entropy loss. This modulating factor decreases the loss contribution from easy examples, allowing the model to focus more on hard examples that are difficult to classify. It is defined as follows:

\begin{equation}
    L_{focal} = -\alpha_t (1 - p_t)^\gamma \log(p_t)
\end{equation}
where $\alpha_t$ is a balancing factor that weights the classes, $\gamma$ is a focusing parameter that adjusts the rate at which easy examples are down-weighted and $p_t$ is the model's estimated probability for the true class. It is calculated using $p$ which is the output of a sigmoid function applied to the network's output logits $x$, $p = \sigma(x)$.

\begin{equation}
    p_t = 
\begin{cases} 
p & \text{if } y = 1 \\
1 - p & \text{if } y = 0 
\end{cases}
\end{equation}
where $y$ in the true class label. In our model we use $\gamma = 2.0$ and $\alpha = 0.25$. 

\subsubsection{Boundary Loss}
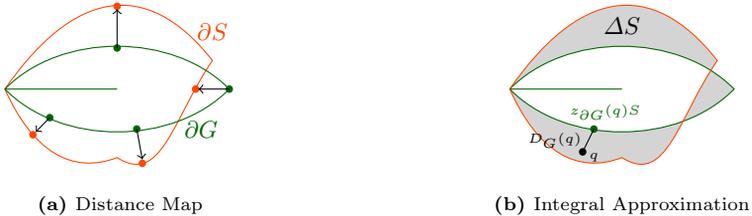
\begin{figure}[t]
  \centering
  \begin{subfigure}{0.45\linewidth}
    \centering
         \begin{tikzpicture}[scale=0.75]
            \definecolor{outlinecolor}{HTML}{FF4500}  
            \definecolor{arrowcolor}{HTML}{000000}
            \definecolor{fillcolor}{HTML}{ADD8E6}
            \definecolor{white}{HTML}{ffffff}
            \draw[outlinecolor] (0,0) .. controls (1.2,1.7) and (2.5,2) .. (3.7,0.5)
                           .. controls (3.0,-0.5) and (2.8,-1.7) .. (2,-1.2)
                           .. controls (1.2,-1.5) and (0.5,-1.1) .. (0,0);

            \node[outlinecolor] at (3.7,1) {$\partial S$};
            
            \draw[draw=bordercolor] (0,0) .. controls (1,1) and (3,1) .. (4,0)
                              .. controls (3,-1) and (1,-1) .. (0,0);

            \node[bordercolor] at (3.5,-0.7) {$\partial G$};

            \draw[draw=bordercolor] (0,0) -- (2,0);
            \coordinate (B) at (2,0.72);
            \coordinate (C) at (4,0);
            \coordinate (D) at (2.35,-0.7);
            \coordinate (E) at (0.8,-0.5);
            
            \coordinate (B') at (2.0,1.45);
            \coordinate (C') at (3.4,0.0);
            \coordinate (D') at (2.45,-1.3);
            \coordinate (E') at (0.5,-0.8);

            \foreach \coord in {B, C, D, E} {
                \node[circle, fill=bordercolor, inner sep=1pt] at (\coord) {};
            }
            
            \foreach \coord in {B', C', D', E'} {
                \node[circle, fill=outlinecolor, inner sep=1pt] at (\coord) {};
            }

            \coordinate (B'') at (2.0,1.4);
            \coordinate (C'') at (3.45,0.0);
            \coordinate (D'') at (2.45,-1.23);
            \coordinate (E'') at (0.55,-0.75);
            
            \draw[arrowcolor,->] (B) -- (B'');
            \draw[arrowcolor,->] (C) -- (C'');
            \draw[arrowcolor,->] (D) -- (D'');
            \draw[arrowcolor,->] (E) -- (E'');
            
    \end{tikzpicture}
    \caption{Distance Map}
    \label{fig:dist}
  \end{subfigure}
  \hfill
  \begin{subfigure}{0.45\linewidth}
    \centering
    \begin{tikzpicture}[scale=0.75]
            \definecolor{outlinecolor}{HTML}{FF4500}  
            \definecolor{arrowcolor}{HTML}{000000}
            \definecolor{fillcolor}{HTML}{d4d4d4}
            \definecolor{white}{HTML}{ffffff}
            \definecolor{black}{HTML}{000000}

            \begin{scope}
                \clip 
                    (0,0) .. controls (1,1) and (3,1) .. (4,0)
                    .. controls (3,-1) and (1,-1) .. (0,0);
                
                \fill[fillcolor, even odd rule] 
                    (0,0) .. controls (1.2,1.7) and (2.5,2) .. (3.7,0.5)
                    .. controls (3.0,-0.5) and (2.8,-1.7) .. (2,-1.2)
                    .. controls (1.2,-1.5) and (0.5,-1.1) .. (0,0)
                    -- (0,0) .. controls (1,1) and (3,1) .. (4,0)
                    .. controls (3,-1) and (1,-1) .. (0,0);
            \end{scope}
        
            \begin{scope}
                \clip 
                    (0,0) .. controls (1.2,1.7) and (2.5,2) .. (3.7,0.5)
                    .. controls (3.0,-0.5) and (2.8,-1.7) .. (2,-1.2)
                    .. controls (1.2,-1.5) and (0.5,-1.1) .. (0,0);
                
                \fill[fillcolor, even odd rule] 
                    (0,0) .. controls (1,1) and (3,1) .. (4,0)
                    .. controls (3,-1) and (1,-1) .. (0,0)
                    -- (0,0) .. controls (1.2,1.7) and (2.5,2) .. (3.7,0.5)
                    .. controls (3.0,-0.5) and (2.8,-1.7) .. (2,-1.2)
                    .. controls (1.2,-1.5) and (0.5,-1.1) .. (0,0);
            \end{scope}

            \draw[draw=bordercolor] (0,0) .. controls (1,1) and (3,1) .. (4,0)
                              .. controls (3,-1) and (1,-1) .. (0,0);

            \draw[outlinecolor] (0,0) .. controls (1.2,1.7) and (2.5,2) .. (3.7,0.5)
                           .. controls (3.0,-0.5) and (2.8,-1.7) .. (2,-1.2)
                           .. controls (1.2,-1.5) and (0.5,-1.1) .. (0,0);

            \draw[draw=bordercolor] (0,0) -- (2,0);

            \node[] at (2.0,1.1) {$\Delta S$};
            \node[bordercolor, font=\tiny] at (1.7,-0.4) {$z_{\partial G}(q) S$};
            \node[circle, fill=bordercolor, inner sep=1pt] at (1.5,-0.7) {};
            \node[circle, fill=black, inner sep=1pt] at (1.3,-1.1) {};
            \node[font=\tiny] at (0.8,-0.9) {$D_{G}(q)$};
            \node[font=\tiny] at (1.5,-1.2) {$q$};

            \draw[black,-] (1.5,-0.7) -- (1.3,-1.1);
            
    \end{tikzpicture}
    \caption{Integral Approximation}
    \label{fig:integral}
  \end{subfigure}
  \caption{The boundary of ground truth area $G$, $\partial G$, is shown in green and the boundary of segmentation mask $S$, $\partial S$ is shown in orange.}
  \label{fig:short}
\end{figure}

To further mitigate the challenges of highly unbalanced segmentation problems, we add a boundary loss function. We use the function proposed by Kervadec et al. \cite{kervadec2019boundary} which is based on integrals over the interface between regions instead of unbalanced integrals over the regions. Figure \ref{fig:dist} shows the desired distance calculation $Dist(\partial G, \partial S)$, where the $L_2$ distance is calculated from each point of the boundary of G, $\partial G$, to the corresponding point on the boundary of S, $\partial S$, along the direction normal to the boundary of $G$. Since these point-wise differentials would be both difficult to compute and difficult integrate into a loss function directly, Kervadec et al. propose an integral approximation, illustrated in \ref{fig:integral}:

\begin{equation}
    Dist(\partial G, \partial S) \approx 2 \int_{\Delta S} D_G(q)dq
\end{equation}
where $\Delta S$ is the region between the contours. $D_G : \Omega \to \mathbb{R}$ is a distance map with respect to the boundary $\partial G$ which evaluates the distance between point $q$ and the nearest point $z_{\partial G}$ on the contour $\partial G$: $D_G(q) = \| q - z_{\partial G} \|$ .

Then a pre-computed level set function $\phi_G : \Omega \to \mathbb{R}$, derived from $D_G(q)$, is used to represent the boundary $\partial G$:

\begin{equation}
    \phi_G(q) = 
\begin{cases} 
-D_G(q) & \text{if } q \in G \\
D_G(q) & \text{otherwise}
\end{cases}
\end{equation}

It is possible to incorporate an approximation of this distance function as the following loss function:

\begin{equation}
    L_{boundary} = \alpha \int_{\Omega} \phi_G(q) s(q) dq
\end{equation}
where $\alpha$ is a weight that increases throughout training. We initially set $\alpha = 0.01$ and increase it by 0.0006 at each epoch. $s(q)$ is a function that returns the probability $p$ for each position $q$.

\subsubsection{Loss Function}
While in the original Mask2Former model dice and cross-entropy loss is used to calculate mask loss. Our implementation replaces cross-entropy loss with focal loss and add boundary loss:

\begin{equation}
    L_{\text{mask}} = L_{\text{focal}} + L_{\text{dice}} + L_{\text{boundary}}
\end{equation}

The mask loss and class loss is calculated for both the plants and leaves,  respectively. The total loss is calculated as a weighted sum of the classification and mask loss of the plants and leaves:

\begin{equation}
    L = \lambda_{\text{cls}} L^{p}_{\text{cls}} + \lambda_{\text{mask}} L^{p}_{\text{mask}} + \lambda_{\text{cls}} L^{l}_{\text{cls}} + \lambda_{\text{mask}} L^{l}_{\text{mask}}
\end{equation}
where $L^p$ and $L^l$ are the losses for the plants and leaves respectively. The weights for each of the losses were set to $\lambda_{\text{mask}}=2.5$ and $\lambda_{\text{cls}}=1.0$.

To facilitate better convergence, deep supervision is used so that the loss function is applied at each layer in the transformer decoders.

\subsection{Metrics}

\subsubsection{Standard Panoptic Segmentation Metrics}
As in \cite{weyler2022field}, panoptic quality (PQ) \cite{kirillov2019panoptic} is used to assess the predicted masks of crops \(PQ_{crop}\) and leaves \(PQ_{leaf}\). The average over these values is reported as $PQ$. During evaluation, plant or leaf instances where less than 50\% of its pixels are within the image, do not affect the score, since these are regarded as uninformative. Additionally, the IoU is calculated for the ``stuff'' categories: weeds \(IoU_{weed}\) and soil \(IoU_{soil}\). The metric \(PQ^\dagger\) is the average over \(PQ_{crop}\), \(PQ_{leaf}\), \(IoU_{weed}\) and \(IoU_{soil}\).

\subsubsection{Leaf Counting Metrics}
In addition to these standard metrics we also analyzed the accuracy of the leaf counts. For this we use root mean square (RMSE) defined as:

\begin{equation}
    \text{RMSE} = \sqrt{\frac{1}{n} \sum_{i=1}^{n} (x_i - \hat{x_i})^2}
\end{equation}

To calculate the error, predicted crop masks and ground truth crop masks are matched according to pairs with the highest IoU. If there is no corresponding instance with an IoU greater 50\% then the instance is considered unmatched. 

To understand the influence of unmatched crop instances on the leaf counting error we calculate the error in three ways. First only over matched crop instances, then over all predicted crop instances, and finally over all ground truth crop instances. These metrics are defined as follows:
\begin{itemize}
    \item $\mathbf{RMSE_{TP}}$ evaluates the leaf counting error with $n$ equal to all true positive crop predictions. These are predicted crop masks that were matched with a ground truth crop mask. This shows the counting error without accounting for the role of unmatched crop instances in the predictions or the ground truth.
    \item $\mathbf{RMSE_{Pred}}$ evaluates the leaf counting error with $n$ equal to all crop predictions. Where a crop prediction could not be matched with a ground truth crop, the ground truth leaf count is 0. This accounts for the role of false positive crop predictions in leaf counting error.
    \item $\mathbf{RMSE_{GT}}$ evaluates the leaf counting error with $n$ equal to all crops in the ground truth. Where a ground truth crop could not be matched with a prediction, the predicted leaf count is 0. This shows the counting error accounting for the role of false positive crop predictions.
\end{itemize}

\section{EXPERIMENTS}
\subsection{Dataset}
Our approach was tested against the PhenoBench dataset \cite{weyler2023dataset}. Examples of the dataset images can be seen in Figure~\ref{fig:examples}. It consists of RGB images of sugar beet crops and weeds taken from a UAV (Figure~\ref{fig:examples} (a)). These images were annotated on three levels: first plants, weeds, and soil were semantically segmented, second plant (crop and weed) instances were segmented (Figure~\ref{fig:examples} (b)), and finally each leaf instance of the sugar beet crops was segmented (Figure~\ref{fig:examples} (c)). The training set contains 1407 images, the validation set contains 772 images and the test set contains 693 images. The images have a resolution of $1024\times 1024$. Further details about the dataset collection and annotation process can be found in \cite{weyler2023dataset}.

\subsection{Training Settings}
Due to the limited training data available, we used a model pretrained on COCO from \cite{cheng2021mask2former}. We trained the model with two different backbones: ResNet-50 \cite{he2015deep} and SwinL \cite{liu2021swin}. The former is less compute intensive and therefore more applicable for real-world use cases where compute is constrained. The latter, though compute intensive, is a more powerful feature extractor that enables our model to achieve performance comparable with the state-of-the-art. We use Detectron2 \cite{wu2019detectron2} and, as suggested for Mask2Former \cite{cheng2021mask2former}, we use the AdamW \cite{loshchilov2019decoupled} optimizer and the step learning rate schedule. We use an initial learning rate of 0.0001 and a weight decay of 0.05. A learning rate multiplier of 0.1 is applied to the backbone and we decay the learning rate at 0.9 and 0.95 fractions of the total number of training steps by a factor of 10. The batch size was 4. The model was trained for 200 epochs.

The best-performing benchmark on the Phenobench dataset used a random sample of the validation set in training \cite{cvppa}, so their training set had 1792 images. For comparison purposes, in addition to experiments on the standard dataset, we show our results on a training set that includes a random sample of the validation set. However, this comparison is not perfect, as our random sample of the validation set may differ. When trained with this larger dataset, we observed the model's performance continued to improve when trained for 300 epochs, unlike the models trained on the smaller dataset, which did not show similar progress. Therefore, the results for this model are based on training for 300 epochs. To facilitate easy comparison with all other benchmarks, unless otherwise specified, our results are presented based on a model trained with the standard dataset of 1407 training images for 200 epochs.

\subsection{Data Augmentation}
At training time we use data augmentation that randomly flips images horizontally and randomly rotates them by 90\textdegree, 180\textdegree, or 270\textdegree. We also tried test time augmentation (TTA) with the same set of augmentations at test time, but only found any benefit when we consider the average results on leaf segmentation. As a result, the only time TTA is used in our experiments is on the leaf segmentation.

\section{Results}
\begin{figure}[t]
\centering
\begin{subfigure}{0.3\columnwidth}
    \centering
    \includegraphics[width=0.95\linewidth]{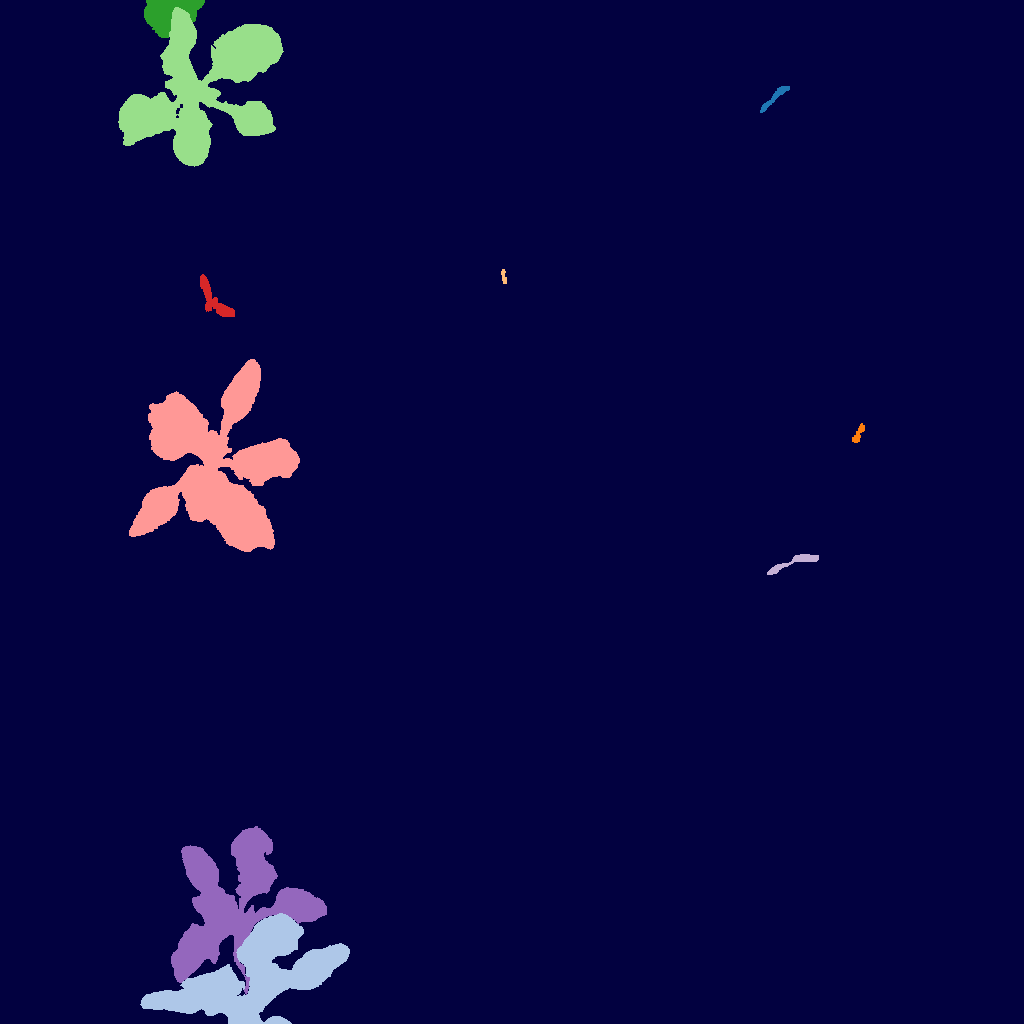}\vspace{0.3cm}
    \includegraphics[width=0.95\linewidth]{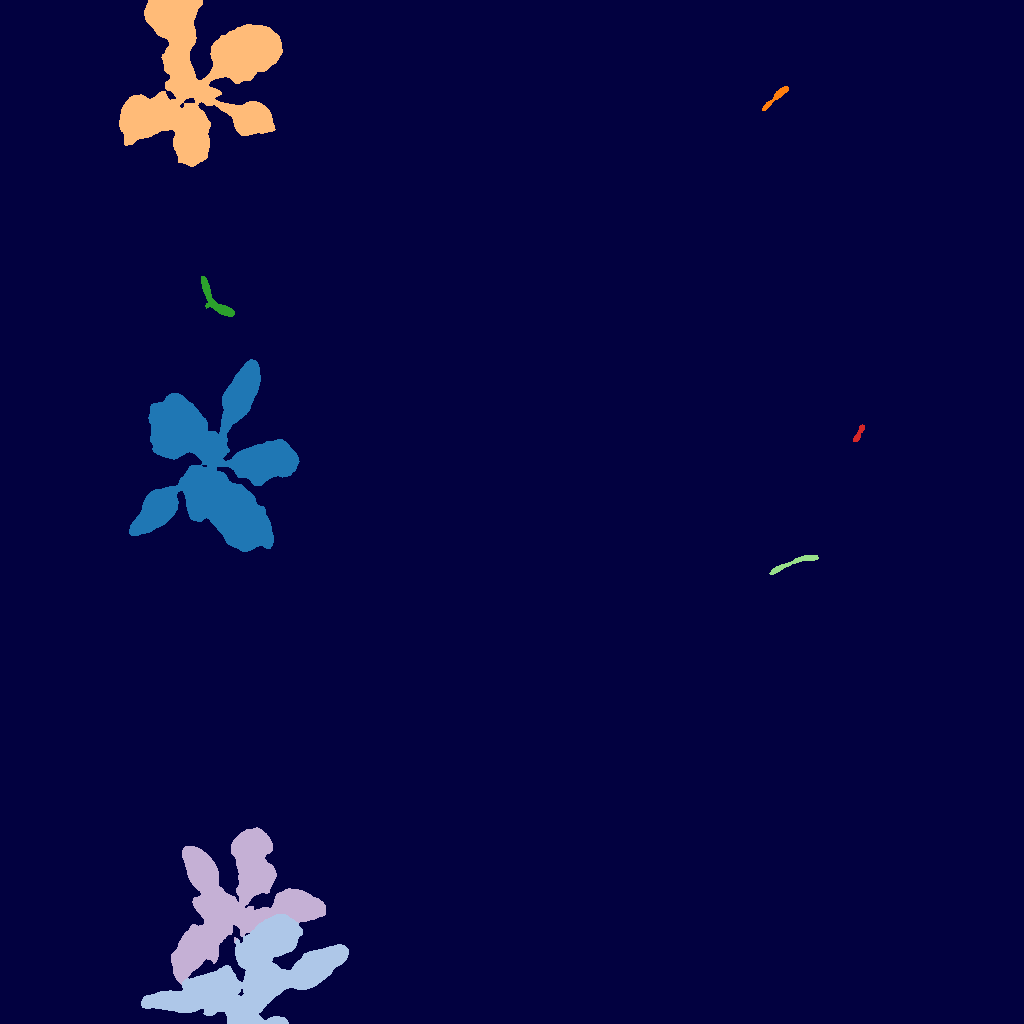}
    \caption{Plant}
    \label{fig:results:plant}
\end{subfigure}
\begin{subfigure}{0.3\columnwidth}
    \centering
    \includegraphics[width=0.95\linewidth]{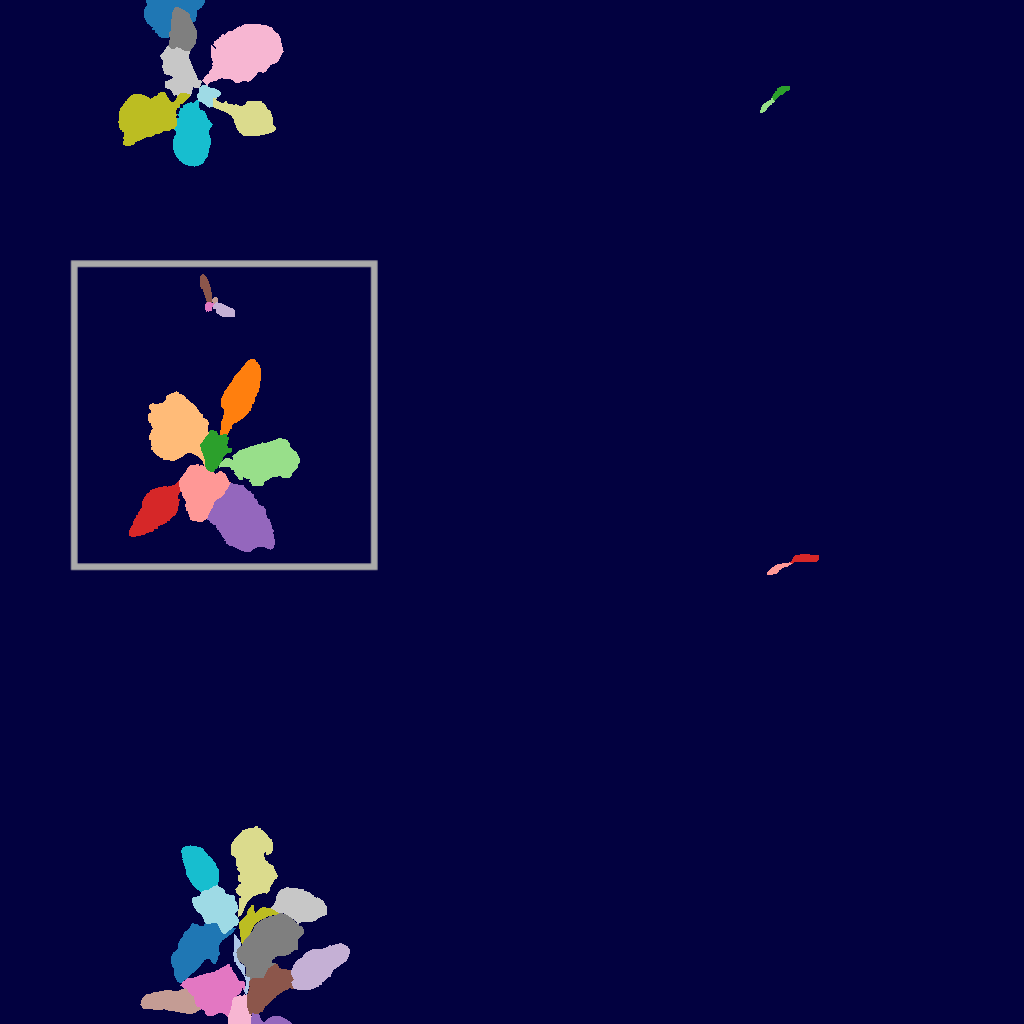}\vspace{0.3cm}
    \includegraphics[width=0.95\linewidth]{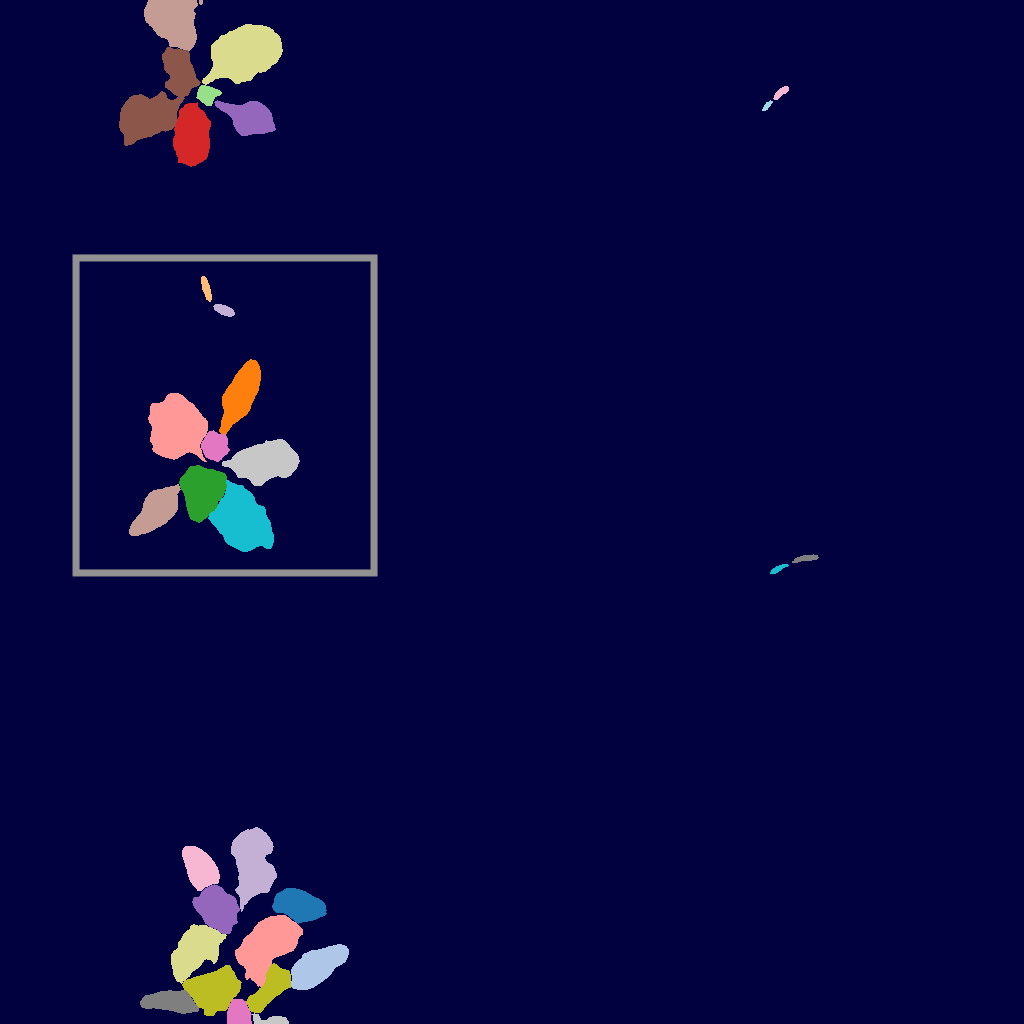}
    \caption{Leaf}
    \label{fig:results:leaf}
\end{subfigure}
\begin{subfigure}{0.3\columnwidth}
    \centering
    \includegraphics[width=0.95\linewidth]{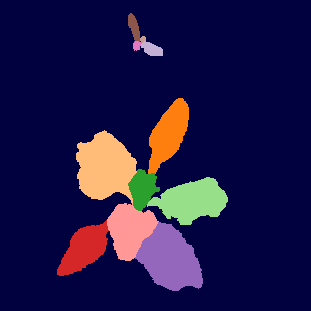}\vspace{0.3cm}
    \includegraphics[width=0.95\linewidth]{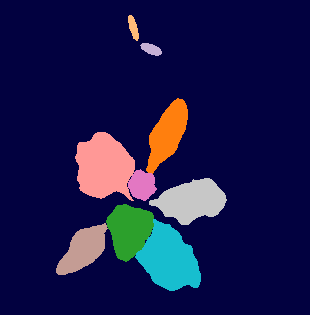}
    \caption{Leaf Zoomed}
    \label{fig:results:zoomed}
\end{subfigure}
\caption{Comparison of ground truth segmentations (top) with the segmentation results on SwinL ($L_f+L_b$) + TTA (bottom). The grey square in \ref{fig:results:leaf} demarcates the zoomed area shown in \ref{fig:results:zoomed}.}
\label{fig:results}
\end{figure}

Table \ref{tab:ablations} shows the ablations of our model with a ResNet-50 backbone and demonstrates the loss functions, and TTA on standard panoptic segmentation metrics. Table \ref{tab:swin} shows the performance of our proposed model with the more powerful SwinL backbone. Table \ref{tab:counts} shows a comparison of leaf counting performance for each of our models. Figure \ref{fig:results} shows an example of segmentation from the validation set. The top row is the ground truth and the bottom row is our predictions. Table \ref{tab:comparison} shows a comparisons of our model with existing benchmarks.

The variations in the loss function we evaluate do not affect the inference speed of the model. Therefore, without TTA and using an Nvidia A100 with a batch size of 4, each model with a ResNet-50 backbone achieved 23.75 frames per second and the model with a SwinL backbone achieved 5.62 frames per second.

\begin{table}[t]
  \caption{Ablations on the validation set using standard metrics.}
  \label{tab:ablations}
  \centering
  \begin{tabular}{p{3.8cm}p{1cm}p{1cm}p{1cm}p{1cm}p{1cm}p{1cm}p{0.8cm}p{0.8cm}}
    \toprule
    Model & $IoU_{soil}$ & $IoU_{crop}$ & $IoU_{weed}$ & $PQ_{crop}$ & $PQ_{leaf}$ & $PQ_{weed}$ & $PQ$ & $PQ^{\dagger}$ \\ \midrule
    ResNet50 (Base) & \textbf{99.45} & 95.74 & 72.94 & 81.86 & 71.68 & 49.17 & 76.77 & 81.48 \\
    ResNet50 ($L_f$) & 99.44 & 95.72 & 72.62 & 81.66 & 72.12 &  50.02 &  76.89 & 81.46 \\
    ResNet50 ($L_b$) & \textbf{99.45} & \textbf{95.78} & 73.64 & \textbf{81.96} & 72.06 & 51.16 & 77.01 & 81.78 \\
    ResNet50 ($L_f+L_b$) &  \textbf{99.45} & 95.76 & \textbf{73.75} & 81.75 & 72.29 & \textbf{51.56} & 77.02 & 81.81 \\ 
    ResNet50 ($L_f+L_b$)+TTA & \textbf{99.45} & 95.76 & \textbf{73.75} & 81.75 & \textbf{72.61} & \textbf{51.56} & \textbf{77.18} & \textbf{81.89} \\
  \bottomrule
  \end{tabular}
\end{table}

\begin{table}[t]
  \caption{Performance of the proposed model using a SwinL backbone.}
  \label{tab:swin}
  \centering
  \begin{tabular}{p{3.8cm}p{1cm}p{1cm}p{1cm}p{1cm}p{1cm}p{1cm}p{0.8cm}p{0.8cm}}
    \toprule
    Model & $IoU_{soil}$ & $IoU_{crop}$ & $IoU_{weed}$ & $PQ_{crop}$ & $PQ_{leaf}$ & $PQ_{weed}$ & $PQ$ & $PQ^{\dagger}$ \\ \midrule
    SwinL ($L_f+L_b$) & \textbf{99.48} & \textbf{96.09} & \textbf{76.39} & \textbf{83.7} & 74.92 & 55.61 & 79.31 & 83.62 \\
    SwinL ($L_f+L_b$) + TTA & \textbf{99.48} & \textbf{96.09} & \textbf{76.39} & \textbf{83.7} & \textbf{75.24} & \textbf{55.61} & \textbf{79.47} & \textbf{83.70} \\
  \bottomrule
  \end{tabular}
\end{table}


\begin{table}[tb]
  \caption{Ablations on the validation set using leaf counting metrics. The precision and recall of crop masks is given. A positive crop mask is defined as a one with best matching ground truth mask with an IoU of over 50\%.}
  \label{tab:counts}
  \centering
  \begin{tabular}{p{3.8cm}p{1.3cm}p{1.2cm}p{1.6cm}p{1.6cm}p{1.6cm}}
    \toprule
    Model & Precision & Recall & $RMSE_{TP}$ & $RMSE_{Pred}$ & $RMSE_{GT}$ \\ \midrule
    ResNet50 (Base) & \textbf{0.76} & 0.94 & 1.61 & 2.11 & 2.35 \\
    ResNet50 ($L_f$) & 0.75 & 0.94 & 1.60 & 2.10 & 2.34 \\
    ResNet50 ($L_b$) & 0.76 & 0.94 & 1.55 & 2.08 & 2.31 \\
    ResNet50 ($L_f+L_b$) & 0.75 & 0.94 & 1.52 & 2.05 & 2.30 \\ 
    ResNet50 ($L_f+L_b$) + TTA & 0.75 & 0.94 & 1.58 & 2.06 & 2.30 \\ 
    SwinL ($L_f+L_b$) + TTA & 0.75 & \textbf{0.95} & \textbf{1.37} & \textbf{1.99} & \textbf{2.24}   \\
  \bottomrule
  \end{tabular}
\end{table}

\begin{table}[tb]
  \caption{Comparisons of our model with other published approaches and competition submissions tested on the private test set. *The 1st place submission randomly sampled half of the validation set to train their model. To provide some comparison, we did the same, but there are likely discrepancies in the samples that may impact the results. The performance of these models should be compared separately from the others so they are highlighted in blue.}
  \label{tab:comparison}
  \centering
  \begin{tabular}{@{}lllllllll@{}}
    \toprule
    Approach & \#Training Images & $IoU_{soil}$ & $IoU_{weed}$ & $PQ_{crop}$ & $PQ_{leaf}$ & $PQ$ & $PQ^{\dagger}$ \\ \midrule
    Weyler et al. \cite{weyler2022field} & 1407 & - & - & 38.37 & 42.60 & 40.43 &  -     \\
    HAPT \cite{roggiolani2023hierarchical}      & 1407 & 98.50 &  61.11 & 54.61 & 46.84 & 50.73 & 65.27 \\
    3rd Place \cite{cvppa} & 1407 & 99.35 & 70.1 & 81.82 & 72.98 & 77.4 & 81.06 \\
    2nd Place \cite{cvppa} & 1407 & 99.18 & 70.66 & \textbf{81.66} & \textbf{73.81} & 77.73 & 81.33 \\
    \mfd{1st Place} \cite{cvppa} & \mfd{1792*} & \mfd{\textbf{99.44}} & \mfd{\textbf{74.13}} & \mfd{82.04} & \mfd{\textbf{74.86}} & - & \mfd{\textbf{82.62}} \\ \midrule
    SwinL ($L_f+L_b$) & 1407 & \textbf{99.41} & \textbf{72.9} & 81.62 & 73.64 & \textbf{77.63} & \textbf{81.89} \\
    \mfd{SwinL ($L_f+L_b$)} & \mfd{1792*} & \mfd{99.42} & \mfd{72.76} & \mfd{\textbf{82.31}} & \mfd{74.15} & \mfd{78.23} & \mfd{82.16} \\
  \bottomrule
  \end{tabular}
\end{table}

\section{Discussion}
\subsubsection{Ablations}
The results in Table \ref{tab:ablations} demonstrate the power of Mask2Former as a segmentation approach because even with a ResNet50 backbone and no enhancements a \(PQ^\dagger\) of 81.48 is achieved. Our results show the benefit of both focal loss and boundary loss to enhance the accuracy of leaf and weed segmentation. This does come at the expense of a slight drop in $PQ_{crop}$ but because it boosts other metrics more it leads to an overall improvement in $PQ^{\dagger}$. This is promising because it shows that even a fast, lightweight model can achieve strong performance with our proposed refinements.

\subsubsection{Leaf Counting}
The focus on improving leaf segmentation is further supported when evaluating leaf counting error, as shown in Table \ref{tab:counts}. The precision and recall of crop masks remain fairly static despite variations in $PQ_{crop}$. It appears improvements in the $RSME_{Pred}$ and $RSME_{GT}$ correlate more with improvements in leaf segmentation than improvements in crop segmentation. 
It should be noted that reductions in leaf counting are fairly modest across all models. Additionally, improvements in panoptic quality do not always translate into a reduction in leaf count. A further analysis of the factors that lead to leaf counting error may highlight new strategies for reducing error that are not so focused solely on improving segmentation performance.

\subsubsection{Qualitative Analysis}
Figure \ref{fig:results} shows an example segmentation from the validation set. Particularly with bigger plants, the segmentation seems pretty accurate for both plants and leaves. However, the small pale orange weed plant, shown in the ground truth, is missing from the prediction in Figure \ref{fig:results:plant}. Additionally, Figure \ref{fig:results:zoomed} shows some of the inaccuracies of our method in more detail. For example, while the leaf count of the larger plant is correct, the shape of the leaves could benefit from additional refinement. Furthermore, our method only segments two leaves in the small plant above where the ground truth includes the two small additional leaves, revealing the true leaf count is 4. This illustrates the challenge of counting leaves correctly and the necessity to tune models, as we have tried to, toward better segmenting very small areas.

\subsubsection{Comparison with Other Methods}
Table \ref{tab:comparison} shows the comparison of our approach with existing published approaches. The 2nd place submission from Nguyen et al. \cite{cvppa} which used a model based on SAM \cite{kirillov2023segment} is the best performing model on the standard training set, achieving a $PQ^{\dagger}$ of 81.33. Our model outperformed this, achieving a $PQ^{\dagger}$ 81.89, however, we had a marginally lower $PQ_{crop}$ and $PQ_{leaf}$. 

The first place submission from Lu et al. \cite{cvppa} used a random sample of the validation set in training so they used a training set of size 1792. We did the same for comparison purposes, however, our random sample of the validation set may have differed so it's not a perfect comparison. Despite this, their implementation does outperform ours, achieving a $PQ^{\dagger}$ 82.62 compared to our 82.16, with only $PQ_{crop}$ being lower at 82.04 compared to our 82.31. Their approach is Mask2Former-based as well but they used a BEiT-L \cite{bao2021beit} backbone rather than SwinL \cite{liu2021swin}. In future work, we will evaluate the performance of additional backbone architectures and evaluate the trade-off between segmentation performance, inference speed, and memory usage.

Overall, despite these shortcomings, we believe our work still has value because the loss function we employed improves the model's performance without affecting the speed or model footprint at inference time. For real-world usage, a more lightweight model may be necessary, and the strategies we have employed here will still be relevant in these instances. 

\section{Conclusion}
In conclusion, in this work, we have investigated the use of several techniques that can enhance segmentation performance, and tested them on a crop-weed segmentation dataset. Our findings highlight that these techniques---in particular focal loss, boundary loss and the use of a separate transformer decoder for leaves---can improve performance on relatively lightweight architectures, making them viable for real-world applications. Furthermore, we have assessed the accuracy of leaf counts calculated from our segmentation and demonstrated that our approach is competitive with the state-of-the-art.

\bibliographystyle{splncs04}
\bibliography{main}
\end{document}